\documentclass[letterpaper, 10 pt, conference]{ieeeconf} 
\usepackage[utf8]{inputenc}
\usepackage[style=ieee,doi=false,isbn=false,url=false]{biblatex}
\usepackage{xcolor}
\usepackage{graphicx}
\usepackage{subcaption}
\usepackage{lipsum}  
\usepackage[normalem]{ulem}
\usepackage{tikz}
\usetikzlibrary{patterns,positioning,arrows,arrows.meta,calc,shapes,pgfplots.groupplots}
\usepackage{pgfplots}
\pgfplotsset{compat=1.9, table/search path={data}}
\usepgfplotslibrary{fillbetween}
\usepackage{xspace}
\usetikzlibrary{chains,decorations.pathreplacing}

\usepackage{import}

\usepackage{siunitx}
\usepackage{booktabs}

\usepackage{multirow, makecell}

\usepackage{balance}

\usepackage{rotating}

\newcommand\scalebarandzoomclip[6]{ 
    \resizebox{\linewidth}{!}{%
        \begin{tikzpicture}[x=#1, y=#1, font=\footnotesize]
            \node[anchor=south west,inner sep=0] (image) at (0,0) {\includegraphics[width=#1,trim={1cm 2.5cm 3cm 5cm},clip]{#2}};
            \coordinate (viewport lower left) at (#3);
            \coordinate (viewport upper right) at (#4);
            \draw[red] (viewport lower left) rectangle (viewport upper right);
            \node[below=0.4cm of image, draw=red, inner sep=0.1pt] (zoom) {\includegraphics[width=#5]{#6}};
            \draw[red, dashed] (viewport lower left) -- (zoom.north west);
            \draw[red, dashed] (viewport upper right|-viewport lower left) -- (zoom.north east);
        \end{tikzpicture}
    }
}

\IEEEoverridecommandlockouts                              % This command is only needed if 
\title{\LARGE \bf
  Semantic 3D Grid Maps for Autonomous Driving 
}

\author{Ajinkya Khoche$^{1}$ $^{2}$, Maciej K Wozniak$^{1}$, Daniel Duberg$^{1}$ and Patric Jensfelt$^{1}$ %<-this % stops a space
\thanks{$^{1}$KTH Royal Institute of Technology, Stockholm 10044, Sweden.
{\tt\small $\{$khoche,maciejw,dduberg,patric$\}$@kth.se}}%
\thanks{$^{2}$The author is employed at Scania CV AB, 151 87 Södertälje, Sweden.
{\tt\small ajinkya.khoche@scania.com}}%
\thanks{This work was supported by PROSENSE (2020-02963) funded by VINNOVA, and by the Wallenberg AI, Autonomous Systems and Software Program (WASP) funded by the Knut and Alice Wallenberg Foundation}%
}

\begin{document}
% \addbibresource{ref.bib} 

\maketitle
\thispagestyle{empty}
\pagestyle{empty}

%%%%%%%%%%%%%%%%%%%%%%%%%%%%%%%%%%%%%%%%%%%%%%%%%%%%%%%%%%%%%%%%%%%%%%%%%%%%%%%%
\begin{abstract}

Maps play a key role in rapidly developing area of autonomous driving.  
We survey the literature for different map representations and find that while the world is three-dimensional, it is common to rely on 2D map representations in order to meet real-time constraints. We believe that high levels of situation awareness require a 3D representation as well as the inclusion of semantic information. We demonstrate that our recently presented hierarchical 3D grid mapping framework UFOMap meets the real-time constraints. Furthermore, we show how it can be used to efficiently support more complex functions such as calculating the occluded parts of space and accumulating the output from a semantic segmentation network.

%Autonomous driving is rapidly developing area of research. We can observe that in the last years many car companies are adding features to their newest models allowing the cars on semi-autonomous driving. While simultaneous localization and mapping (SLAM) plays a key role in autonomous vehicles (AV), usage of the maps is often discarded due to the limitations in computational resources or bandwidth speed. In this article we analyze legacy and state-of-the-art mapping solutions developed in the last years. Moreover we show how detailed 3D maps can be great for different tasks and reveal that the main reason most of the researchers do not use them is lack of an efficient framework. Finally, we show the results and the performance  \textit{UFOMap} which is improved OctoMap framework, that works faster and requires less resources to generate high resolution \textit{voxel} map.    
\end{abstract}

%%%%%%%%%%%%%%%%%%%%%%%%%%%%%%%%%%%%%%%%%%%%%%%%%%%%%%%%%%%%%%%%%%%%%%%%%%%%%%%%
\section{Introduction}
Mapping is essential for autonomous driving. 
A map of the environment can be used to establish situational awareness, localize the autonomous vehicle (AV), plan safe trajectories taking into account the geometry of the road, traffic rules, and the position of surrounding objects~\cite{bresson2017slamsurvey,yurtsever2020survey}.
% It can also enhance the semantic understanding of the surrounding environment~\cite{fei-pillarsegnet}.
Endowing the map with semantic information can further enhance understanding of the surrounding environment~\cite{fei-pillarsegnet}.
%A map can also be used to store sensor readings~\cite{bresson2017slamsurvey}.
If the map is to contribute to the tasks above, the system has to have the ability to analyze the map in \emph{real-time}. This is not trivial to achieve. All data structures and operations to store and access the map must be lightweight and computationally efficient.
%This usually means that instead of storing all sensor readings of the environment, only the ones that are necessary to complete downstream tasks are retained. 
Several modern AV systems have \emph{pre-built} maps on-board, referred to as HD (high definition) Maps. 
These maps simplify real-time calculations by providing useful priors~\cite{kiran2019realtime}.
However, building and maintaining these maps so that they are up-to-date is expensive and requires complex algorithms and large computational resources~\cite{kim2021hd}. 

Regarding the map format, most mapping frameworks store information in 2D grids or in vectorized form. 
% due to the issues mentioned above. 
This usually entails rejecting or simplifying some information, e.g., casting 3D readings into 2D. This also means that some information needed to analyze the long-term complex interactions between dynamic agents is discarded from the map, and has to be handled through dedicated object detection and tracking pipelines. These often work in isolation from the mapping process, acting only on the last few sensor frames. 
%Additionally, most online mapping algorithms make an implicit assumption that the environment is static~\cite{Cadena2016pastpresentfuture}. 
% To handle dynamic objects, the map needs to be manipulated in some way. 
%Current frameworks do not provide an easy way of handling dynamic objects due to high computation cost of necessary operations.
% In the worse case, static assumption can induce failures in algorithms that depend on the map.
%Static assumption an approach can fail completely or induce errors in the algorithms that depend on the map when dynamic objects appear during the mapping process. 

\begin{figure}
    \centering
    \scalebarandzoomclip{0.8\linewidth}{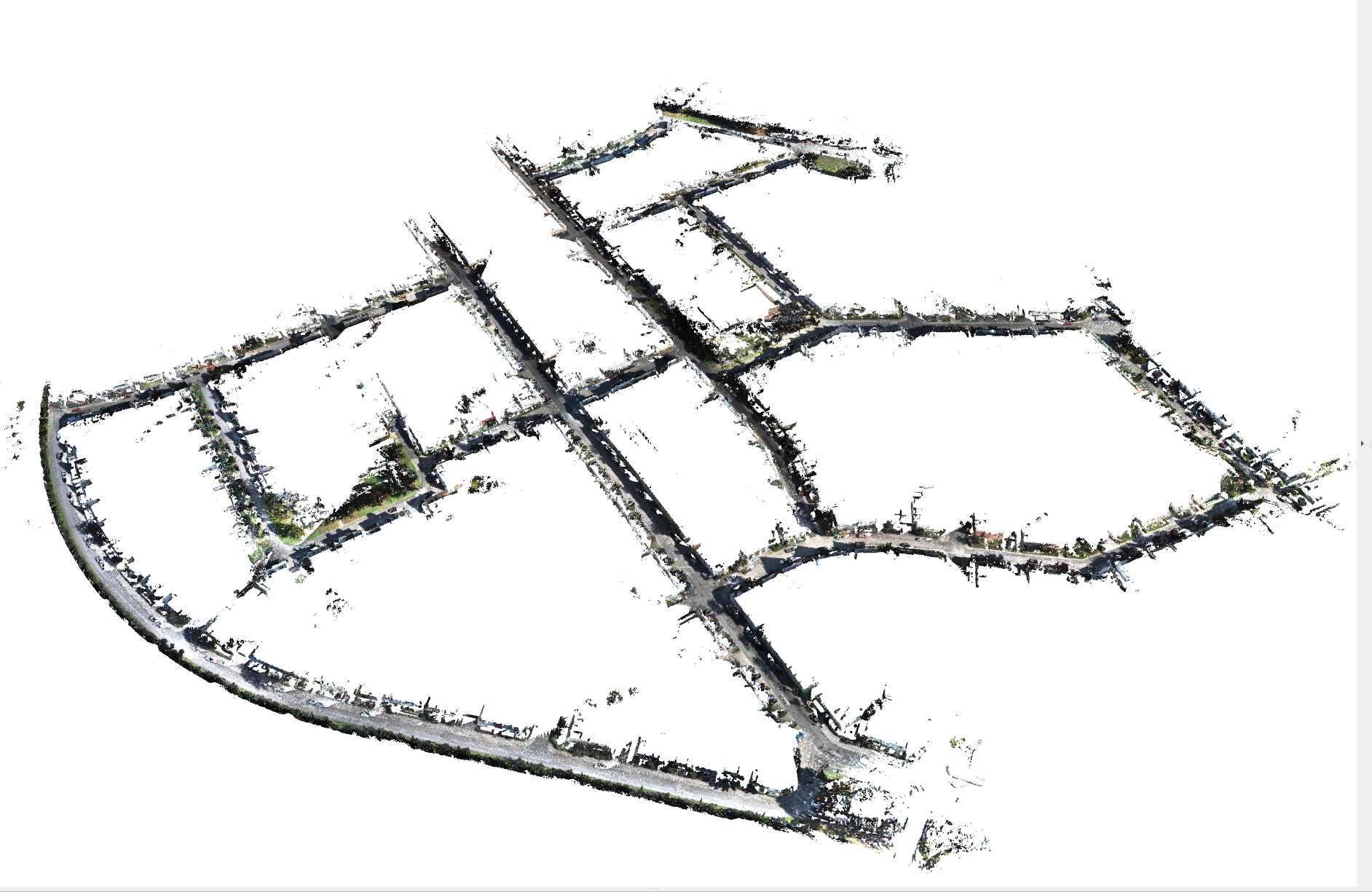}{0.57,0.00}{0.7,0.07}{1.0\linewidth}{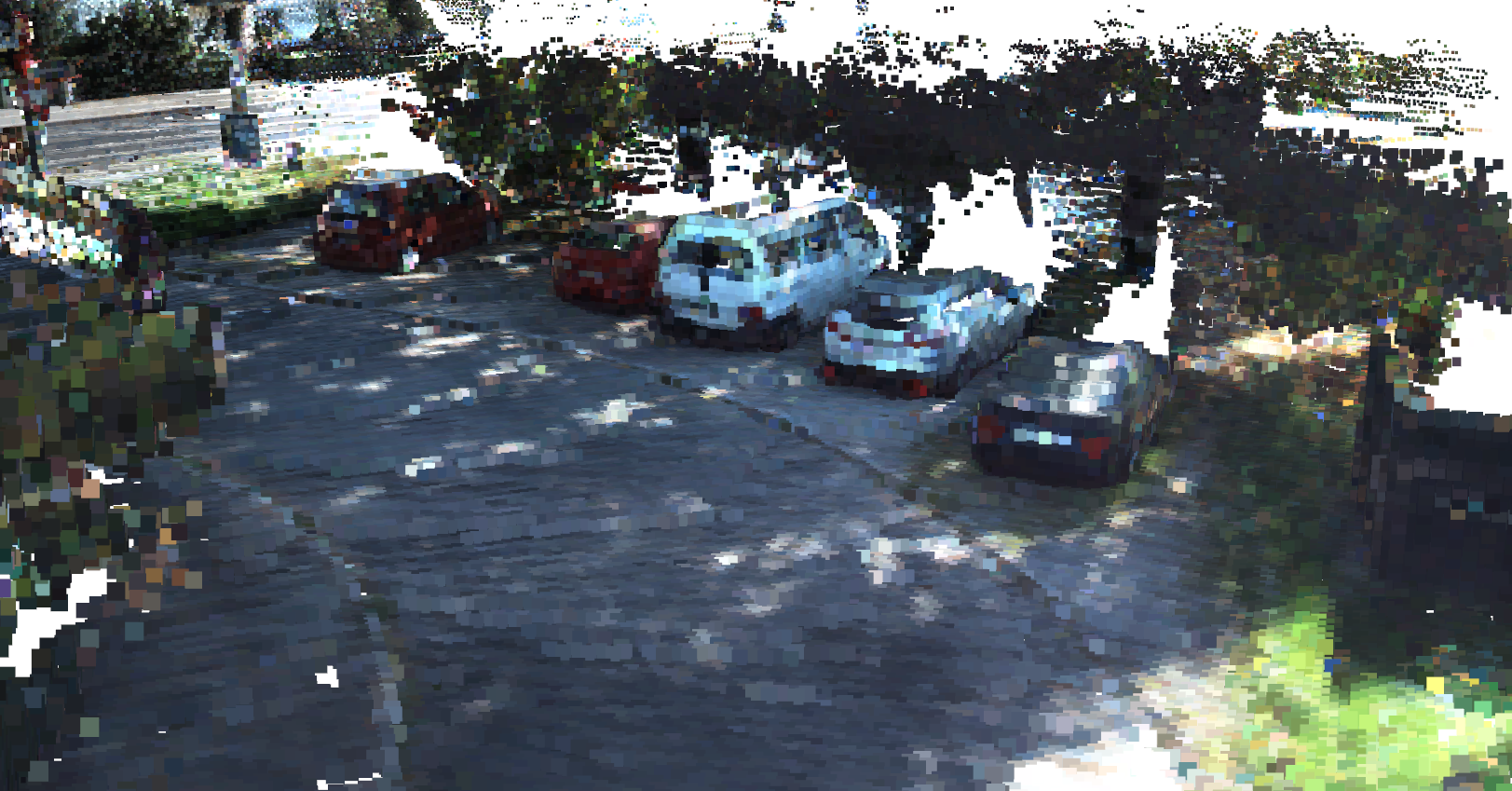}
    \looseness=-1
    \caption{UFOMap with 10 cm voxel size  processing 10 hz data from sequence 00 of SemanticKITII~\protect{\cite{behley2019semantickitti}} dataset in less than 40 ms per sensor frame (single threaded). Top: Full view of the produced map. Bottom: Zoomed in on part of the map.}
    \label{fig:cool_figure}
\end{figure}

% add more clear transition to the DENSE aspect of mapping  --> it can address some issues and these issues were already addressed in the indoor env, but it was not in the oudoor  
% 3d mapping has successfully addressed some of these issues in indoor environments,
%However, this has not extended significantly to outdoor environments
Although 3D semantic mapping algorithms that satisfy real-time constraints are widely used \emph{indoors}, the large scale and complexity of the \emph{outdoor} environment still poses a major challenge. In this paper, we demonstrate that high-resolution dense semantic mapping for AVs is \emph{possible} in real time, using the recently proposed UFOMap~\cite{duberg2020ufomap}. 
%Furthermore, we argue that unknown space (part of the environment that has not been seen) plays a critical role for safety-related reasoning. 
We also believe that a 3D grid may act naturally as an upstream representation from which task-specific representations can be derived.
Furthermore, we show an additional benefit that this representation can provide, by acting as a medium for information fusion. 

%Furthermore, we show that our model enables us to reason about occlusions. 
% Furthermore, we show that this representation can provide a basis for the fusion of information. 
%Additionally, for perception and scene understanding purposes, we argue that being able to store semantic information is extremely important. A occupied cell that is classified as a building is not the same as one classified as a park car. The build will not move, the park car might.

\begin{figure*}
    \centering
    \includegraphics[height=3.04cm]{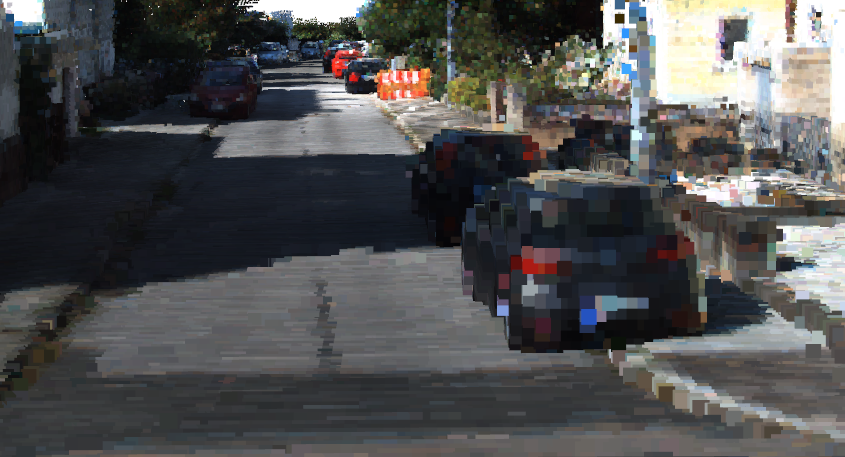}
    \includegraphics[height=3.04cm]{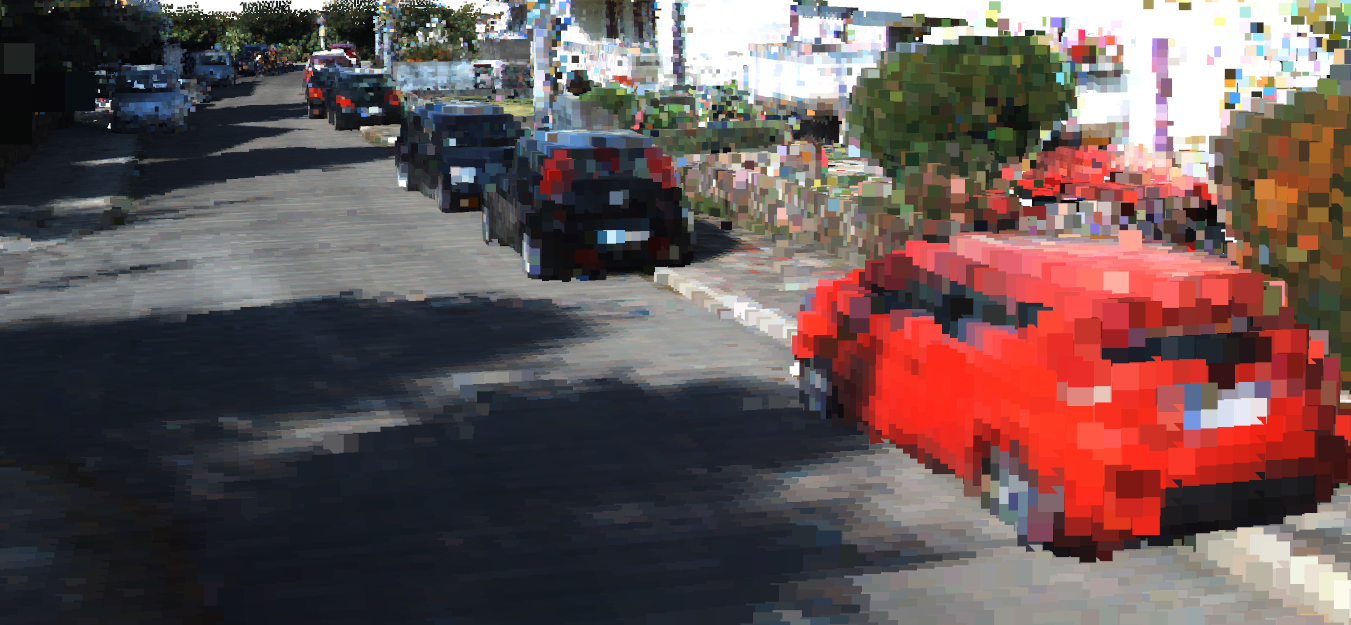}
    \includegraphics[height=3.04cm]{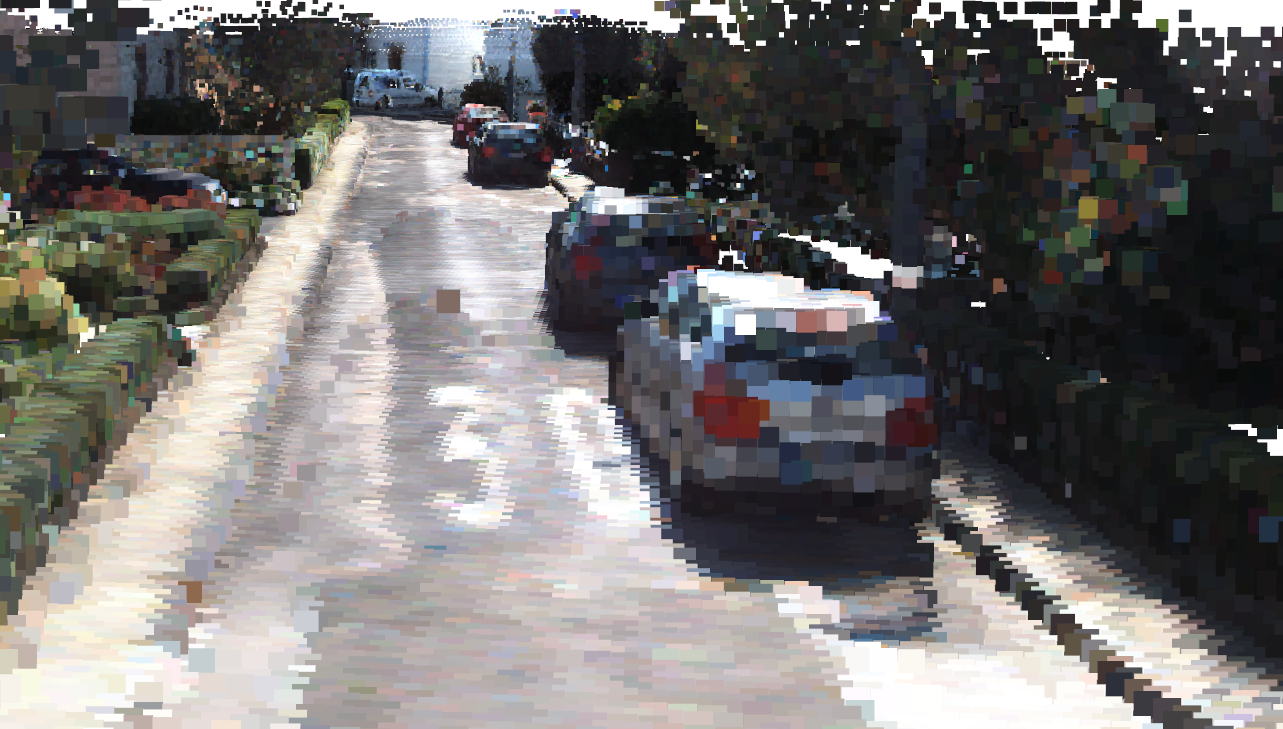}
    
    \vspace{0.2cm}
    
    \includegraphics[height=2.2cm]{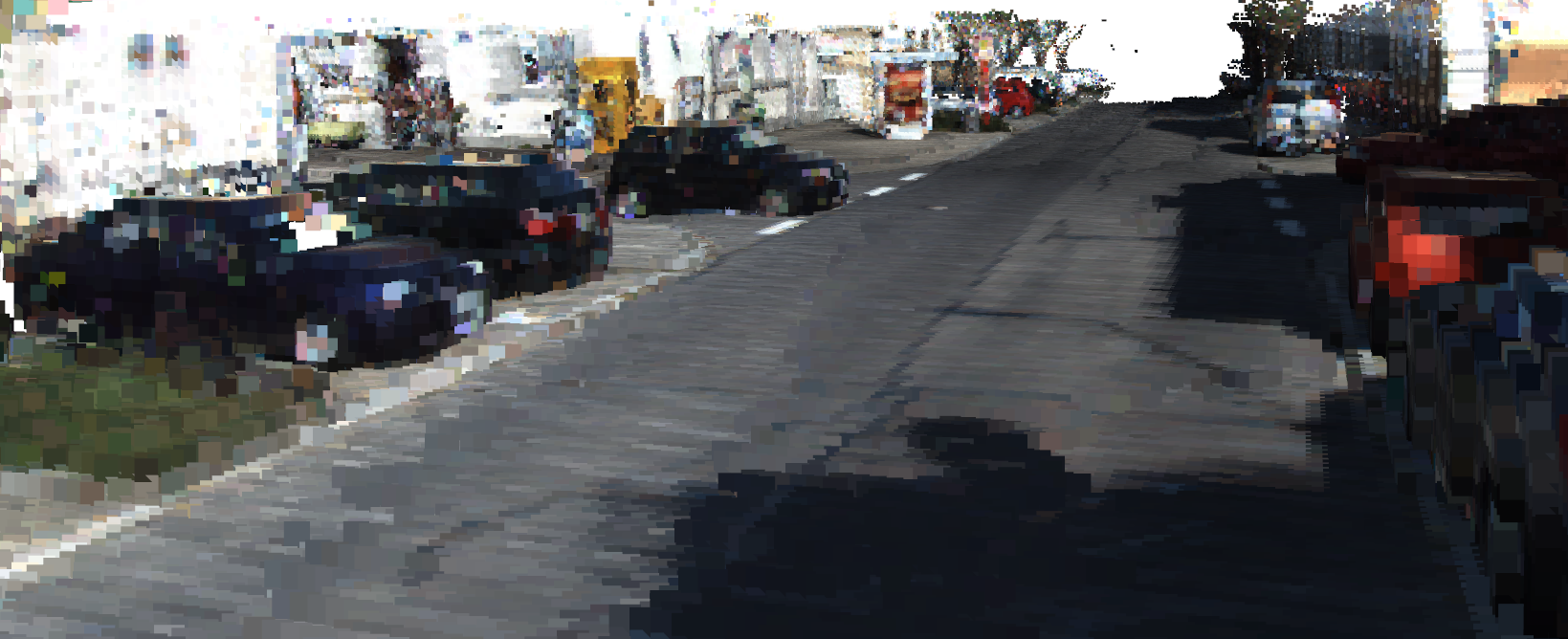}
    \includegraphics[height=2.2cm]{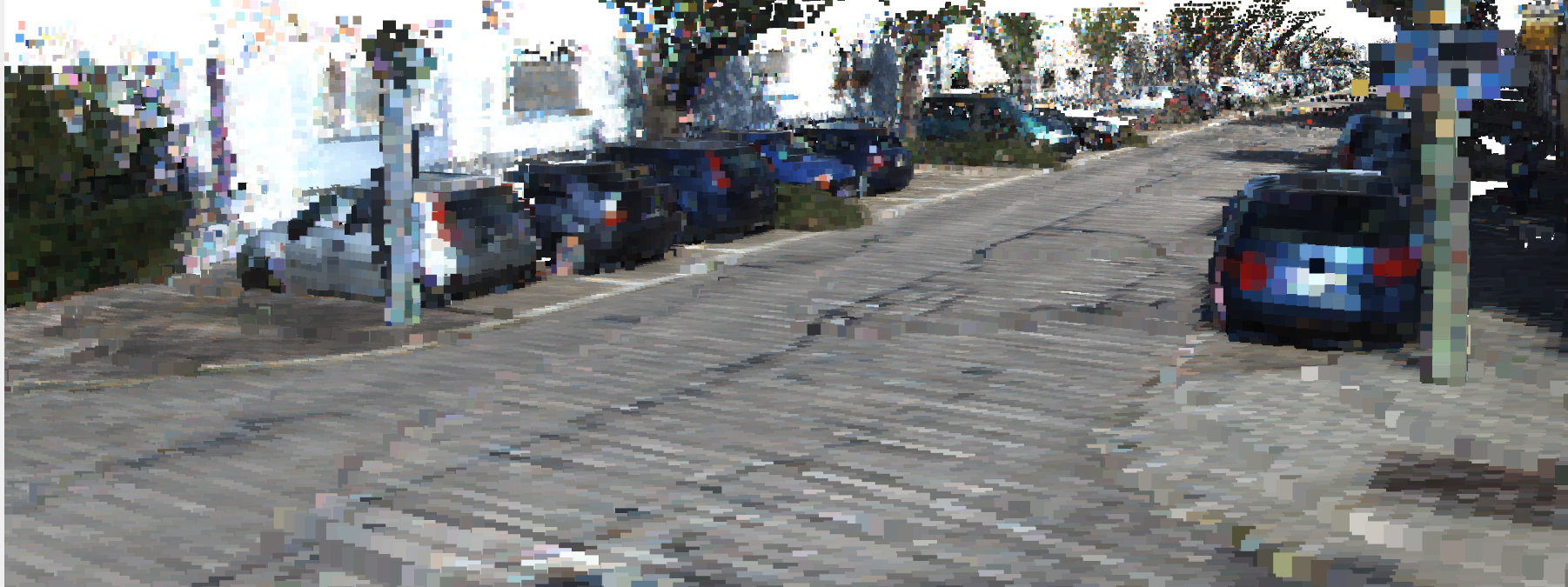}
    \includegraphics[height=2.2cm]{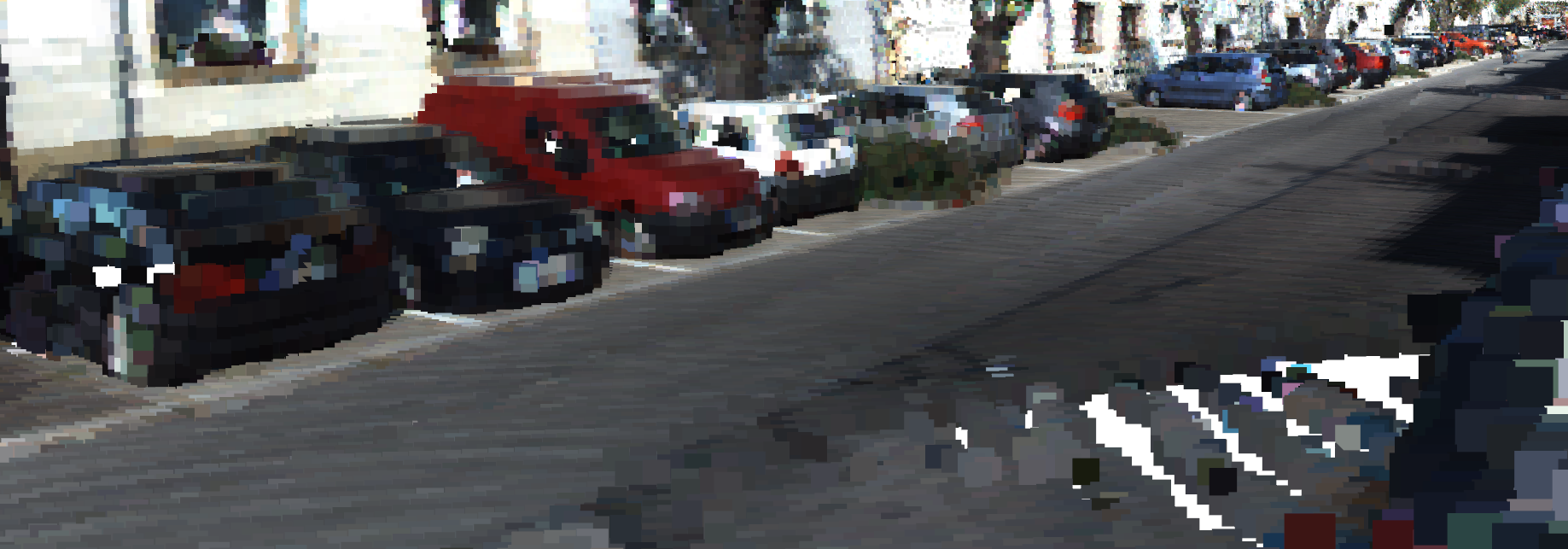}
        
    \caption{Zoomed in on parts of the map shown in Figure~\protect\ref{fig:cool_figure}, demonstrating dense, high-resolution 3D mapping.}
    \label{fig:ufomap_detailed}
\end{figure*}

In this paper, we make the following contributions:
\begin{enumerate}
    \item We survey the literature for map representations in autonomous driving.
    \item We demonstrate that, recent advances in grid mapping allow for dense, high-resolution, real-time 3D mapping in autonomous driving.
    \item We demonstrate how the semantic information can be used to, on-the-fly, manipulate the map information. This would allow support for downstream tasks with different requirements from a common representation.
    % \item efficient generation of 2D projections of the 3D map information so that sensor integration can be done in 3D but for example path planning can be addressed in 2D 
    %%\item We show efficient calculations of the space occluded from the sensors. %% ???
    \item We demonstrate how the map provides the means for sensor / information fusion, exemplified by evaluation on LiDAR based  semantic segmentation.
\end{enumerate}

\section{Map Representations}
\label{sec:map-representations}

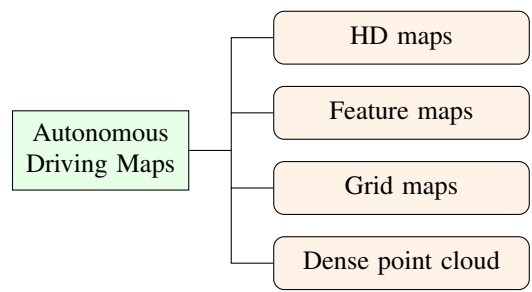
\begin{figure}[!b]
    \centering
    \begin{tikzpicture}[auto,node distance=1cm and 1cm,on grid]%
    % Define block styles
    % \tikzstyle{decision} = [diamond, draw, fill=blue!20, 
    %     text width=4.5em, text badly centered, node distance=3cm, inner sep=0pt]
    \tikzstyle{scenario} = [rectangle, draw, fill=green!10, 
        text width=6em, text centered, minimum height=3em]
    \tikzstyle{map} = [rectangle, draw, fill=orange!10, 
        text width=9em, text centered, rounded corners, minimum height=2em]
    \tikzstyle{line} = [draw]
    \tikzstyle{element} = [ellipse, draw, fill=blue!10, 
        text width=4.2em, text centered, minimum height=2em]
    \tikzstyle{other} = [rectangle, draw, fill=red!10, 
        text width=7em, text centered, rounded corners, minimum height=2em]
    % \tikzstyle{cloud} = [draw, ellipse,fill=red!20, node distance=3cm,
    %     minimum height=2em]
    
    \newcommand*{\connectorH}[4][]{
        \draw[#1] (#3) -| ($(#3) !#2! (#4)$) |- (#4)
    }
    \newcommand*{\connectorV}[4][]{
        \draw[#1] (#3) |- ($(#3) !#2! (#4)$) -| (#4)
    }
    \newcommand*{\connectorHLast}[4][4]{
        \draw[#1] let
                \p1 = (#3),
                \p2 = ($(#3) !#2! (#4)$)
            in (\x2,\y1) |- (#4)
    }
    \newcommand*{\connectorVLast}[4][4]{
        \draw[#1] let
                \p1 = (#3),
                \p2 = ($(#3) !#2! (#4)$)
            in (\x1,\y2) -| (#4)
    }

    % Place nodes
    \node [scenario] (autonomous_driving_maps) {Autonomous Driving Maps};
    
    \node [map, above right=1.5cm and 4cm of autonomous_driving_maps] (hd_maps) {HD maps};
    \node [map, above right=0.5cm and 4cm of autonomous_driving_maps] (feature_based) {Feature maps};
    \node [map, below right=0.5cm and 4cm of autonomous_driving_maps] (grid_maps) {Grid maps};
    \node [map, below right=1.5cm and 4cm of autonomous_driving_maps] (dense_point_cloud) {Dense point cloud};
    
    % \node [element, above right=5mm and 3cm of hd_maps] (vector_map) {Lanelets};
    % \node [element, below right=5mm and 3cm of hd_maps] (rasterized) {Rasterized};

    % \node [element, right=3cm of grid_maps] (elevation) {Elevation};
    % \node [element, above right=10mm and 3cm of grid_maps] (cells) {Cells};
    % \node [element, below right=10mm and 3cm of grid_maps] (voxels) {Voxels};
    
    % Draw edges
    \connectorH [line] {0.50}{autonomous_driving_maps.east}{feature_based.west};
    \connectorHLast [line] {0.50}{autonomous_driving_maps.east}{grid_maps.west};
    % Ensure that two lines do not overlap (just to make it look a bit nicer in PDF viewers with bad AA)
    \path [line] let
            \p1 = ($(autonomous_driving_maps.east) !0.5! (feature_based.west)$),
            \p2 = (feature_based.west)
        in (\x1,\y2) |- (hd_maps.west);
    \path [line] let
            \p1 = ($(autonomous_driving_maps.east) !0.5! (grid_maps.west)$),
            \p2 = (grid_maps.west)
        in (\x1,\y2) |- (dense_point_cloud.west);
    
    % \connectorH [line] {0.50}{hd_maps.east}{vector_map.west};
    % \connectorHLast [line] {0.50}{hd_maps.east}{rasterized.west};
    
    % \connectorHLast [line] {0.50}{grid_maps.east}{cells.west};
    % \path [line] (grid_maps.east) -- (elevation.west);
    % \connectorHLast [line] {0.50}{grid_maps.east}{voxels.west};
\end{tikzpicture}%
    \caption{A non-exhaustive overview of different map representations used in autonomous driving. HD maps are used to store map prior information, such as the location of lanes and drivable areas. Feature maps are commonly used for localization purposes and contain pose-descriptor pairs for landmarks. Grid maps discretize the world and are typically built online from sensor data and used as input for planning. In autonomous driving 2D grids are common, but 3D versions are rare because of the computational cost. Using the point clouds produced by, e.g., a LiDAR as the map representation is also common.}
    \label{fig:map_representations}
\end{figure}

In this section, we provide a brief overview of different types of map representations and the information stored in them. Figure \ref{fig:map_representations} presents a non-exhaustive listing of representations. 
Depending on the view one takes, one can arrive at slightly different partitioning, and some methods fall in-between groups. 
%For example, rasterized HD maps are essentially grid maps, but while grid maps are typically built online from sensors rasterized maps are stored priors. 

Two key requirements for subsystems used in an autonomous driving setting are computational efficiency and robustness. Autonomy requires that frames can be processed at a speed that matches the rate at which sensors generate data (i.e., between 10 and 40 FPS~\cite{yang2019re}). Processed here means going from the sensor reading, the analysis of it, deciding which action to take next, and sending the correct signal to, e.g., a throttle or steering unit. 
% Computational efficiency could also mean the memory consumed by certain operation or the disk space needed to store results offline.

% \subsection{HD maps} 
\subsection{HD maps}
\label{sec:hd_maps}
The requirement for robustness in the mapping process has led to some map information being generated offline and acting as priors during a mission. \emph{HD Maps} are used to model, for example, the geometry and topology of the road (connectivity between the road segments). 
This includes lane boundaries, center lines, successor-predecessor relationships, speed limits, or directions~\cite{Liu2019HighDM}. 

Most of the early work relied on manual annotation~\cite{bender2014lanelets} and only recently this process was automatized~\cite{paz2020probabilistic}.
% Still, vector maps have much higher requirements for accuracy compared to feature maps. 
% Vector maps are used by many car companies to represent the drivable space. It is often based on lane limiting features, such as curbs, lines or shoulder of the road.
% One of the earliest formats proposed for this was OpenDRIVE. It used
OpenDRIVE, one of the first formats proposed to systematize road description for driving simulators, used \emph{clothoids} to model road segments\cite{dupuis2010opendrive}. However, representing connectivity in this way proved to be cumbersome.
Bender \emph{et al.}~\cite{bender2014lanelets} represented the bounds of the left and right lane using polylines, which also allowed them to implicitly represent connectivity. Their format, called \emph{Lanelets}, allowed encoding driving rules (called regulatory elements) in complex scenarios like intersections and merging. Lanelets were further extended to include physical objects (e.g., road markings, crash barriers) and areas with restricted movement (e.g., parking, sidewalks or buildings)~\cite{poggenhans2018lanelet2}. 
Lanelet and its extension are lightweight and have been widely adopted for a variety of tasks, such as reachability analysis~\cite{naumann2019safe}, scenario generation~\cite{queiroz2019geoscenario} and benchmarks for motion planning~\cite{althoff2017commonroad} or interaction~\cite{zhan2019interaction}.
% Rasterized maps are another format often used for prior map information. 
% % Many of the new driving datasets distribute information this way. 
% The \textit{NuScenes} dataset~\cite{nuscenes} encode semantic information regarding drivable areas and sidewalks in the rasterized maps. \textit{Argoverse}~\cite{Argoverse} encodes drivable areas and ground height, both at one meter resolution. Using the ground height map one can, for example, reduce false objects detection from LiDAR data.
% TODO

\subsection{Feature maps}
\label{sec:feature_maps}
%These 
% maps store the location of landmarks or features as they are often referred to, especially when image data is used. Examples of landmarks defined by geometric structures in the world are trees, edges and corners. There is a vast amount of work on image based features or keypoints. Image based keypoints come with a descriptor that allows matching between frames. One of the most frequently used SLAM systems, ORB-SLAM3~\cite{orbslam3} rely on the ORB image feature~\cite{rublee2011orb}. Another similar sparse point map is what emerges as output from a struture-from-motion algorithm, such as COLMAP~\cite{schoenberger2016sfm}. Recently, features are often learned from data~\cite{tang2020kp3d}.
%The map itself could be composed entirely of features, points and their associated feature descriptors, or made of the features themselves. 
% Feature maps have been used since the beginning of mapping. 
Feature maps store a sparse description of the environment in a form of features extracted from sensor data and their location in space. 
The extracted features depend on the sensors and the application targeted. Traditionally, image-based features such as edges and corners have been widely explored. The features are often accompanied with a descriptor that simplifies matching.
%in the context of Simultaneous Localization and Mapping (SLAM). %CITE
% in the context of visual SLAM for localization and 3D scene reconstruction. Examples are 
% to create sparse maps which can be used for localization or 3D scene reconstruction.     
%Some of the examples are Shi-Tomasi~\cite{davison2007monoslam}, FAST~\cite{klein2007parallel} or ORB~\cite{mur2015orb}.
%Other methods operate directly on the image, generating a dense or semi-dense maps~\cite{newcombe2011dtam}\cite{forster2014svo}.  
%Many frameworks store representative sensor samples, often referred to as \textit{keyframes}, along with the map or without it. 
Recently, \emph{Deep Neural Networks} (DNN) are used to extract and learn feature representations~\cite{detone2018superpoint, tang2020kp3d}.
Such approaches can be used for robust re-localization under different lighting or weather conditions~\cite{sarlin2021back}. Feature maps support localization related tasks well, but are ill-suited for most other tasks as they provide a sparse representation of objects/obstacles and do not represent free space at all.

\subsection{Dense point clouds}
\label{sec:pc_maps}
% Another common map representation is to represent the world using the sensor data itself. 
% Another common way is to use the sensor data to represent the world. 
% You can, for example, get a map by acquiring LiDAR point clouds from poses some distance apart and aligning them.
Dense point cloud maps can be obtained by aligning LiDAR point clouds and the corresponding poses.
Many approaches have been developed within the area of LiDAR odometry and mapping~\cite{jonnavithula2021lidar}.
Zhang \emph{et al.} achieve real-time LiDAR odometry and mapping by alternating between a fast scan-to-scan and a slower scan-to-map matching process~\cite{zhang2014loam}. 
% between edge and planar features extracted from point clouds.
% They further improve robustness by a adding visual odometry pipeline as a prior.
% Researchers try to improve
Recent work has been aimed at improving the robustness of this solution by adding visual odometry as a prior~\cite{zhang2015visual}, or matching efficiency by projecting 3D points to range images or bird's eye view~\cite{zheng2021efficient}.
Others choose \emph{surfel-based map} representation due to the ease of rendering~\cite{behley2018efficient}.
% Such a map can be seen as a very dense point feature map. The points clouds are typically not associated with descriptors as in feature maps, but recently it is common to assign color information or semantics to the points. High-performance 3D LiDARs can create incredibly dense and accurate point clouds. 
% Once all the dyn Dense maps can be used for localization, or
% In order to be usable for localization, typically a
However, a major drawback of using dense point cloud representation is the lack of scalability, owing to high memory consumption, often preventing researchers to use it in applications for autonomous driving.

\subsection{Grid maps}
\label{sec:grid_maps}
% Grid maps are used in most mobile robot systems
% , at least in some subsystem. 
Grid maps were proposed by Moravec and Elfes~\cite{moravec1985high} and have been refined over the years. The basic idea is to discretize space into 2D or 3D grid cells. 
% In most autonomous driving applications, a 2D representation is chosen for being less computationally expensive and easier to implement and maintain~\cite{Li_2019_CVPR, badue2021self}.
In autonomous driving applications, it is common to choose a 2D representation as it is less computationally expensive and easier to implement and maintain~\cite{badue2021self}.
2D occupancy grids are widely used to model free/occupied space used by a planning algorithm to compute a safe trajectory. 

A middle ground is a 2.5D representation. An example is the elevation map, modeling the ground height. In~\cite{wolcott2017robust}, each cell of the 2D grid stores a Gaussian mixture model capturing the distribution of data along height at the corresponding position. Another way to represent that are \textit{sticks} presented in~\cite{montani1990sticks}. Here voxels with the same feature value are merged along the height direction. It is thus essentially a 2D grid with ``sticks'' representing the data. \emph{Stixels}~\cite{badino2009stixel} similarly are superpixels, defined by clustering points with the same semantic class label in the columns of an image. They proved to be useful and efficient in both static and dynamic environment stettings~\cite{Hehn2019stixels}.

One relatively common map format under the umbrella of HD maps are rasterized maps, which are essentially \emph{2D grids}. 
The \emph{NuScenes} dataset~\cite{nuscenes} encodes drivable areas and sidewalks at a resolution of 10 px$/$m. \emph{Argoverse}~\cite{Argoverse} provides masks for drivable areas and ground height, both at one meter resolution. Using the ground height map, false object detection from LiDAR data can be significantly reduced.

When extending grids to 3D, the cells are referred to as voxels. Typically, fixed sized voxels are allocated dynamically and organised using the voxel hashing algorithm. Voxblox~\cite{ol2016voxblox} is an example of this, storing the truncated signed distance function (TSDF) for each voxel. However, Voxblox struggles when the environment size increases.
An efficient way to represent 3D information is using an \textit{octree}. An octree recursively partitions the space till a smallest resolution (chosen as a design parameter) is reached.
% , which is a recursive space partitioning tree. To represent a list of points in 3D, the space is sub-divided into eight equally sized 
The inherently hierarchical structure enables performing efficient searches at different resolutions.
By prioritizing memory allocation for space with more information, it also provides an efficient way to grow the map.
OctoMap~\cite{hornung2013octomap} is a widely used 3D mapping framework based on octrees, with applications to planning, exploration and localization tasks. On the other hand, OctoMap is still relatively slow when it comes to updating and accessing the information.
UFOMap~\cite{duberg2020ufomap}, also based on octrees, overcomes these shortcomings in OctoMap while also adding new features, as seen in Section \ref{sec:ufomap}. 
\section{Additional map use cases}
\label{sec:use_cases}
% isn't better to say applications? or soem other name for the section
As seen in previous section, maps are most often aimed at localization, path planning, and decision making. 
%Going beyond the usual use cases, This section lists some advanced applications that maps can enable. 
This section lists some other use cases for maps.

% \section{Semantics in Mapping}
\subsection{Semantic Mapping}
\label{sec:semantics}
% \subsubsection{Semantics}
% Semantic understanding is an intermediate step between sensor input and decision making. 
% Most of the representations covered in~\ref{sec:map-representations} can be enriched with semantic information. 
% However, it is important to understand which information semantic representation can provide.
In computer vision, semantics is the ability to classify background and foreground~\cite{kirillov2019panoptic}.
Semantic segmentation methods using DNN acting on camera and LiDAR data can be used to directly annotate dense maps, feature maps, and grid maps~\cite{garg2020semantics}.
This allows using the same output for multiple tasks, for example, drivable area detection or road landmark detection~\cite{paz2020probabilistic}.
Recently, researchers have also achieved promising results in instance-level tracking~\cite{aygun20214d}.

% The drawback is that the neural networks are only trained on a fixed number of classes predetermined by the producers of information. Any modification to this class list would require expensive re-labelling and re-training.
However, neural networks are only trained on a fixed number of classes predetermined by the authors of the dataset, which can create the following problems. For instance, a certain group might be interested in segmenting different types of vehicle (e.g., car, truck, ambulance, bus, police), while another group might be satisfied with grouping all the vehicles under one label.
One could think of a map as a way to maintain such different views of semantics.

\subsection{Information Fusion}
\label{sec:information_fusion}
% Common sensors in autonomous driving are cameras, LiDARs, and radars. Cameras acquire projections of the 3D world onto the 2D image plane. They are ideal for detecting and classifying objects and general scene understanding. Combining images from two or more cameras or using deep learning methods allow depth to be estimated. LiDAR return point clouds and is excellent at measuring distances. In some cases, intensities are also available, creating a form of LiDAR image. Radars have lower resolution than LiDAR, but are less sensitive to the environment disturbances and can directly measure the speed of moving objects. 

An AV usually fuses information from multiple sensors to achieve robustness. In general, there are two types of fusion processes: late fusion and early fusion. In late fusion, the information is first processed at the sensor level and then fused. On the other hand, for early fusion, the sensor data is first fused together and then processed. 
% , depending on the order in which the information is interpreted
% An example of late fusion in the context of object detection would be that detection is done with each sensor itself first, and then bounding boxes are added. In early fusion, we fuse the raw data, for example, by projecting the point cloud onto the image and adding depth information.

Sensor fusion can also occur over time by combining the readings acquired at different times and positions. 
Consider again the example of using DNN to estimate semantic segmentation. Researchers often limit themselves to using single images or LiDAR scans as input. Observing that there is a sufficient overlap between measurements, it should be reasonable to expect that fusing successive estimates of the neural network, as the vehicle moves forward, should improve overall semantic segmentation. Although researchers have investigated this idea, they often limit applications to indoor environments~\cite{Rosinol20icra-Kimera, narita2019panopticfusion, grinvald2019volumetric} or 2D representations of outdoor environments~\cite{paz2020probabilistic, Bieder2020ExploitingMG, Fei2021PillarSegNetPS}.

Given that the world is 3D, projecting things down to 2D causes information loss and creates a simplified map that makes it difficult, for example, to track temporally occluded objects or estimate the 3D pose of moving objects~\cite{xiang2017subcategory,8621614,xiang2015data}. We argue that a 3D grid is ideal for information fusion.
%The mapping framework we propose allows us to do so in real-time. 
Furthermore, a 3D model of the world would provide a better model for occlusions and also allow, for instance a better understanding of pedestrians' behavior~\cite{5765690} or off-road terrain~\cite{6629540,maturana2018real}.

\section{Limitation of current methods} %frameworks ?
\label{sec:lim}
The map representations described in Section~\ref{sec:map-representations} may be robust or computationally efficient but are designed around specific applications, making it hard to use them for any other task. To enable advanced vehicle autonomy, we believe that map representations should satisfy two more requirements: \emph{flexibility} and \emph{ease of use}. 

Maps cater to a variety of downstream tasks, the nature and structure of which can change over time. Flexibility refers to the ability of the mapping framework to adapt to these changes. Failure to do so limits their usability, or worse, risks future development being constrained by the mapping framework design. In order to be easy to use, the mapping framework should enable fast access to information of interest. It should also provide users an intuitive way to query for said information. To our knowledge, none of the existing mapping frameworks for autonomous driving fulfill these requirements, while also being dense, 3D, and satisfying real-time constraints. In the experiments, we show that UFOMap~\cite{duberg2020ufomap} can fulfill these requirements.

\section{UFOMap Mapping Framework}
\label{sec:ufomap}
In this work, we build on the UFOMap 3D grid mapping framework\footnote{\url{https://github.com/UnknownFreeOccupied/ufomap}}. In the following, we provide a brief overview for convenience. UFOMap uses the octree data structure. The voxel at the highest resolution is called the \emph{leaf node}, and is used to store occupancy. The sensor data is integrated into the map using ray-tracing, i.e. a ray is projected from the sensor origin and the probability of occupancy at leaf node corresponding to the measured point is updated,
% leaf node at the measured point is marked as occupied, 
while the other leaf nodes along the ray are marked as free. 
% The probability of occupancy can be updated according to a user-defined function. 
This update function can be user-defined.
By default, the fusion scheme mentioned in~\cite{hornung2013octomap} is used:

\begin{equation}
    L(n|z_{1:t}) = L(n|z_{1:t-1}) + L(n|z_{t})
\label{eq:occ_update}
\end{equation}
where
\begin{equation}
    L(n|z_{t}) = log \big[\frac{P(n|z_{t})}{1-P(n|z_{t})}\big] 
\label{eq:log_odds}
\end{equation}

here $P(n|z_{t})$ is the occupancy probability for leaf node $n$ given measurement $z_{t}$, derived from the beam-based inverse sensor model, and $L(n|z_{t})$ is its log-odds value. UFOMap derives its name from the ability to explicitly represent the unknown space, alongside free and occupied space. This is achieved by placing two thresholds on the occupancy probability.
Implementing fast integrators and incorporating thread safety allows UFOMap to insert or delete information many folds faster than OctoMap~\cite{duberg2020ufomap}.

In addition to occupancy, UFOMap has the ability to store color, semantics, and the timestep information for each voxel.
The timestep at every leaf node indicates how recently a given voxel was updated, as shown in Figure ~\ref{fig:ufomap_time_step}.
% In particular
The semantics is stored as a dynamically allocated array of label-value pairs. This allows UFOMap to store multiple semantic labels per voxel, as well as gives the user the flexibility to expand the list of possible semantics. The label corresponds to the semantics, and the value is an arbitrary quantity attached to a label. In this work, the value represents the number of times a label has been observed. 
% \sout{In particular, semantic information from new measurements is integrated similarly to~\ref{eq:occ_update}}.
To integrate semantic information, Equation \ref{eq:occ_update} is applied to the top-1 label, $l_\text{top-1}$, of each point in a measurement, which means that the value of the label $l_\text{top-1}$ in the voxel increases while the values of the other labels remain unchanged.

UFOMap can efficiently query for
% allows efficient querying of the
information based on both the spatial location and the content itself. 
For example, one could look for all ``cars'' or ``pedestrians'' within a 100 m radius of the ego vehicle.
These queries can also be made at multiple resolutions. For this, information is propagated so that coarser voxels can summarize information in their children. This propagation function can also be user defined.

\section{Experimental Setup}
In this section, we present experiments to demonstrate dense semantic 3D mapping in real time. We address the limitations described in Section \ref{sec:lim} and show some examples on SemanticKITTI dataset~\cite{behley2019semantickitti}.
% , a large scale dataset for autonomous driving. 
% In our experiments, 
LiDAR point clouds, where each point is enriched with color and top-1 semantic label, are fed to UFOMap at 10 hz using ROS~\cite{quigley2009ros}.
When a new sensor measurement arrives, the probability of occupancy as well as semantics for a leaf node is updated according to Equation \ref{eq:occ_update}.
Figure \ref{fig:cool_figure} shows an example of one of the sequences. The map is built with a resolution of 10 cm. Each integration of sensor data takes on average 40 ms using a single CPU thread. Additional detailed views can be seen in Figure \ref{fig:ufomap_detailed}. 

\subsection{Dataset}
% For the experiments we use SemanticKITTI~\cite{behley2019semantickitti}. 
SemanticKITTI adds point-wise semantic and instance labels to KITTI Odometry~\cite{geiger2012we}, a large scale dataset for autonomous driving. The dataset is sequential, and all scans were recorded using Velodyne HDL-64E LiDAR at 10 hz. 
It contains \num{23201} training and \num{20351} testing scans,
% , \num{4549} million points, 
annotated with \num{25} semantic classes. 

\begin{figure}[t]
    \centering
    \includegraphics[width=\linewidth,trim={1cm 2.5cm 3cm 5cm},clip]{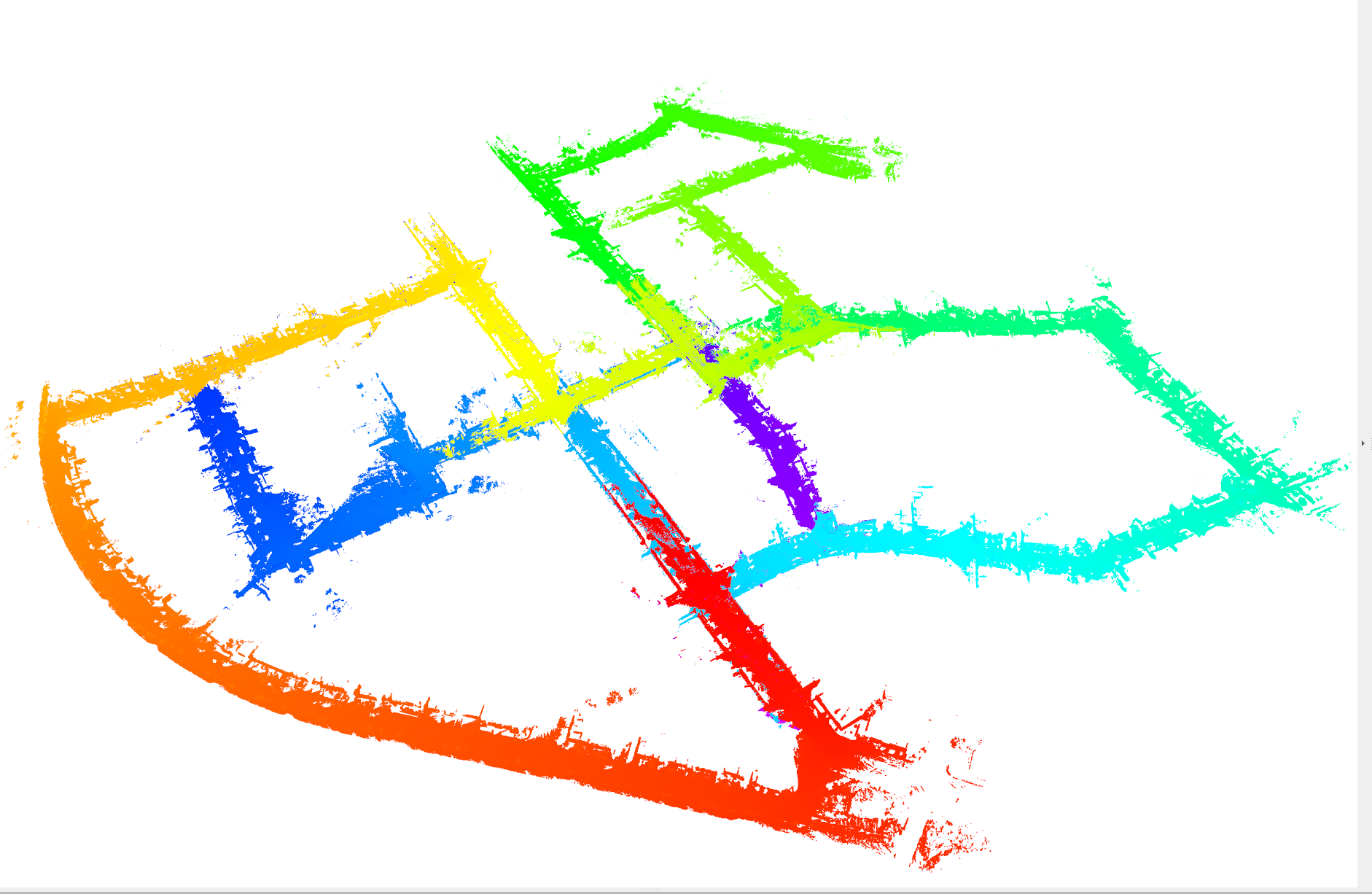}
    \caption{Same as Figure ~\protect\ref{fig:cool_figure}, but showing the time step stored in the voxels. Blue to red color imply older to newer time steps.}
    \label{fig:ufomap_time_step}
\end{figure}

% In our experiments, LiDAR point clouds, enriched with color and semantic labels, are fed to UFOMap at \qty{10}{\hertz} using ROS~\cite{quigley2009ros}.
% % \ref{fig:ufomap_rendering} shows qualitative results at an intersection. 
% When a new sensor measurement arrives, the probability of occupancy for a leaf node (a voxel at the lowest level of the tree) is updated according to the probabilistic fusion scheme described in~\cite{hornung2013octomap}. In particular, such an update is also performed for the semantic labels. Voxels are classified by thresholding on the occupancy probability as unknown, free, or occupied. 

% \ref{fig:cool_figure} shows an example of one of the sequences. The map is built with a resolution of \qty{10}{\centi\metre}. Each integration of sensor data takes on average \qty{40}{\milli\second} using a single CPU thread. Additional detailed views can be seen in~\ref{fig:ufomap_detailed}. \ref{fig:ufomap_time_step} illustrates the time-step indicator information in the map. 

\begin{figure*}[t]
    \centering
    \begin{subfigure}[b]{0.4\textwidth}
        \centering
        \includegraphics[width=\textwidth,trim={5.8cm 1.6cm 4.7cm 1.6cm},clip]{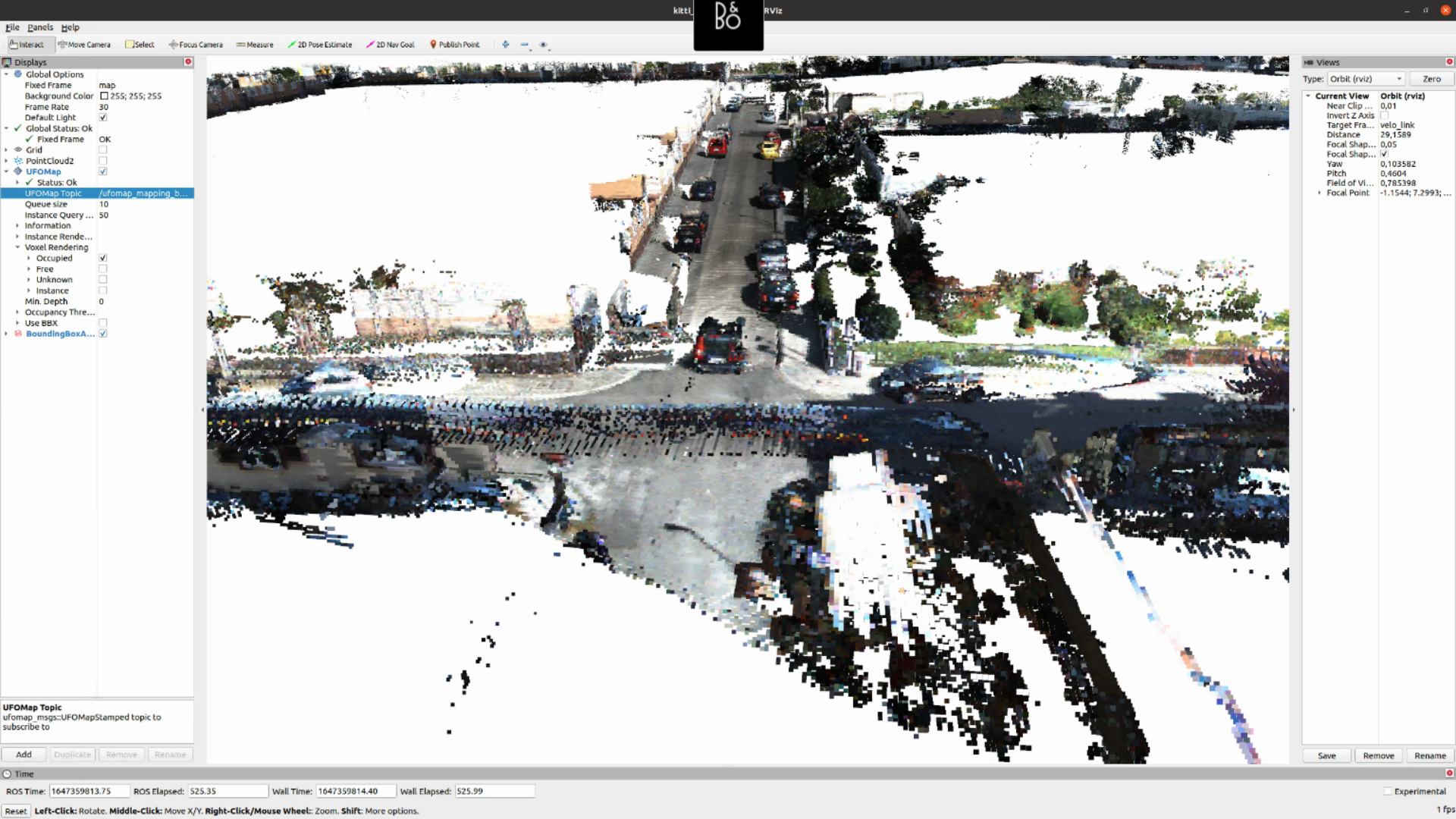}
        \caption{}
    \end{subfigure}
    %\hfill
    \hspace*{0.5cm}
    \begin{subfigure}[b]{0.4\textwidth} 
        \centering
        \includegraphics[width=\textwidth,trim={5.8cm 1.6cm 4.7cm 1.6cm},clip]{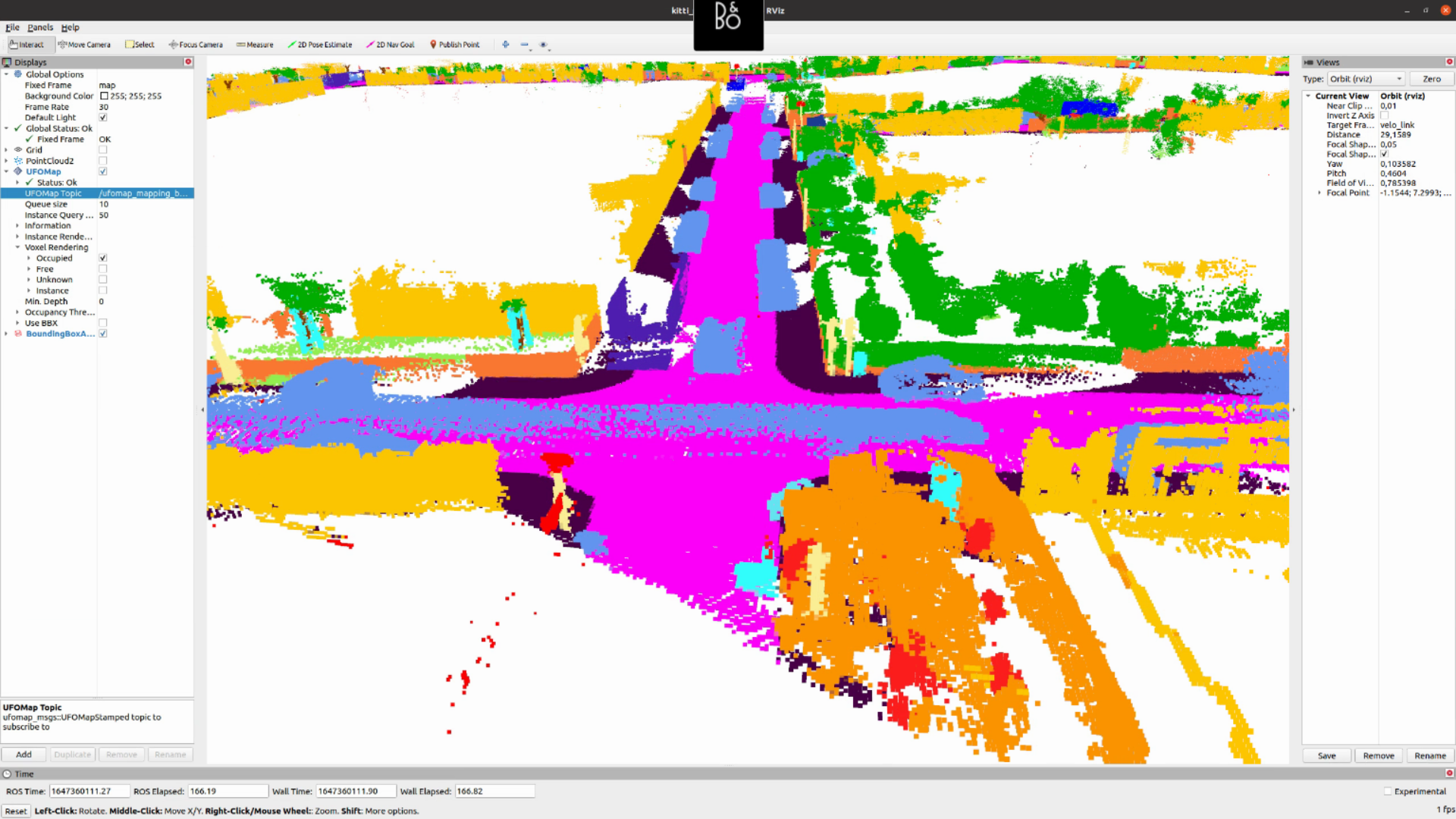}
        \caption{}
    \end{subfigure}
    %\hfill
    %\begin{subfigure}[b]{0.32\textwidth} 
    %    \centering
    %    \includegraphics[width=\textwidth,trim={5.8cm 1.6cm 4.7cm 1.6cm},clip]{figures/semantic_localization_1.PNG}
    %    \caption{}
    %\end{subfigure}
    
    \begin{subfigure}[b]{0.4\textwidth} 
        \centering
        \includegraphics[width=\textwidth,trim={5.8cm 1.6cm 4.7cm 1.6cm},clip]{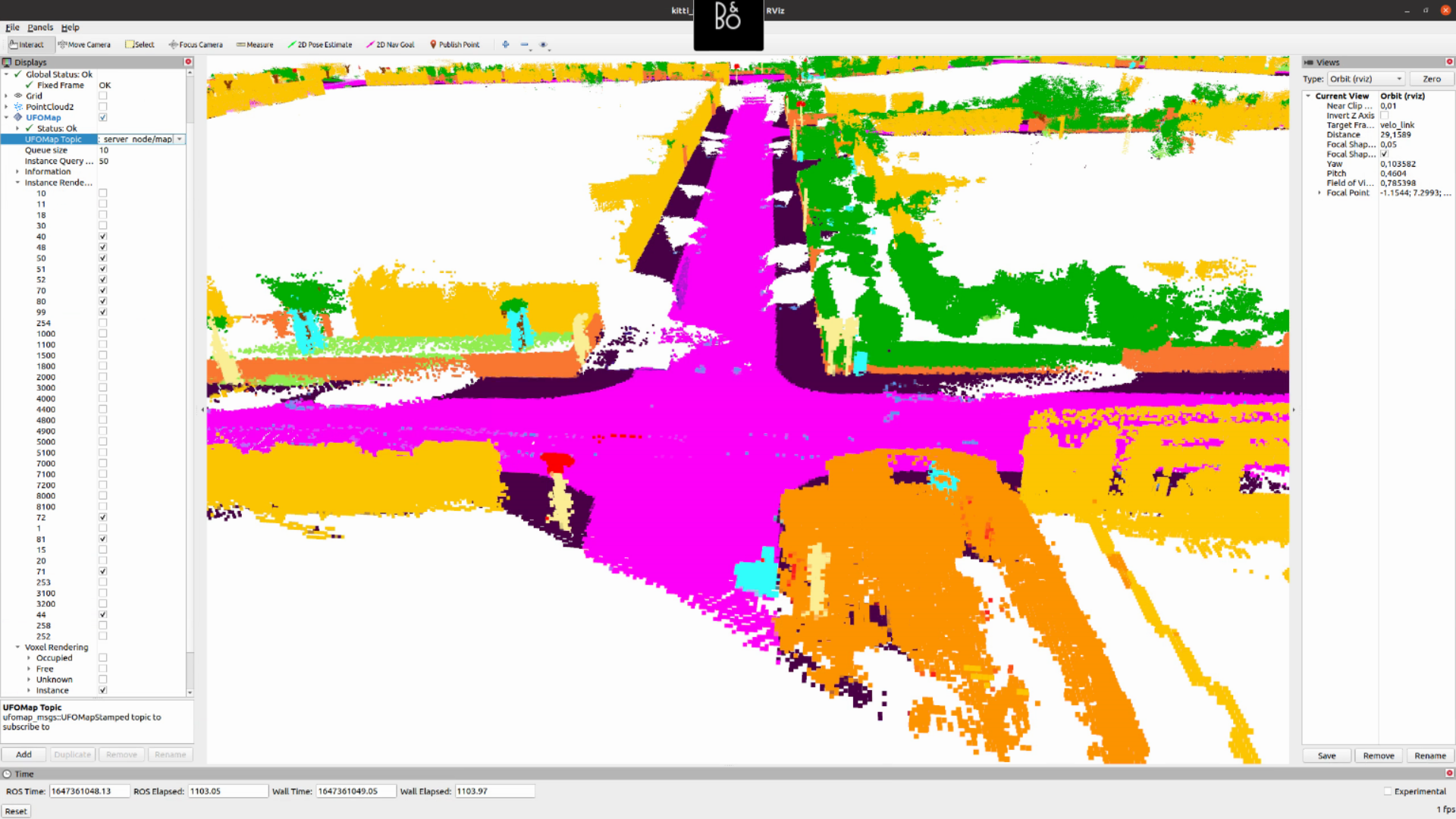}
        \caption{}
    \end{subfigure}
    \hspace*{0.5cm}
    %\hfill
    \begin{subfigure}[b]{0.4\textwidth}
        \centering
        \includegraphics[width=\textwidth,trim={5.8cm 1.6cm 4.7cm 1.6cm},clip]{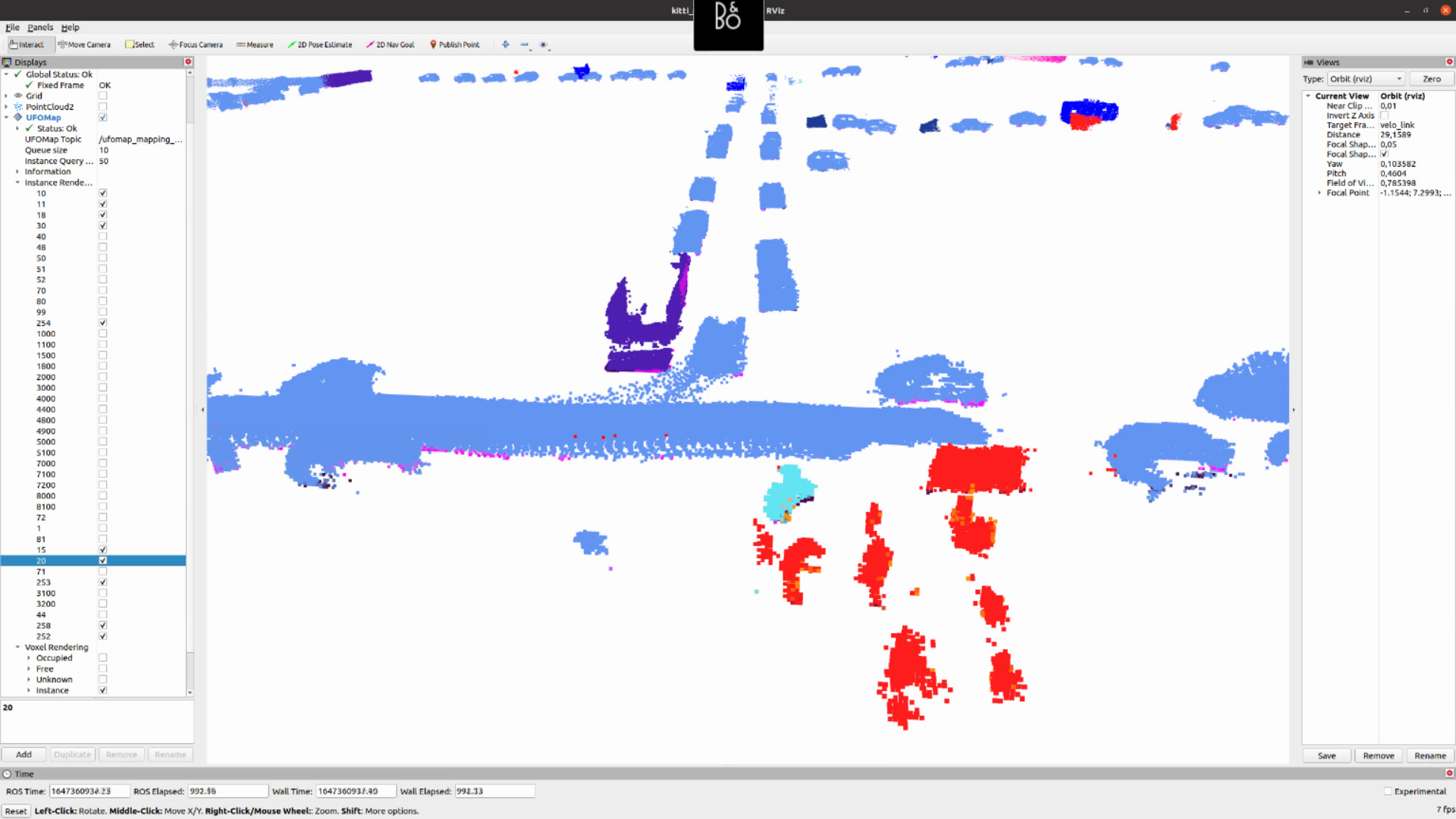}
        \caption{}
    \end{subfigure}
    %\hfill
    %\begin{subfigure}[b]{0.32\textwidth}
    %    \centering
    %    \includegraphics[width=\textwidth,trim={5.8cm 1.6cm 4.7cm 1.6cm},clip]{figures/semantic_motion_1.PNG}
    %    \caption{}
    %\end{subfigure}
    \caption{Rendering occupied voxels from UFOMap at 10 cm resolution on SemanticKITTI~\cite{behley2019semantickitti} dataset, sequence 07. (a) Color and (b) Semantic information respectively for the entire scene. (c) Background and (d) Foreground objects. 
    %Third Column: (c) geometric map for localization and (f) Objects with motion. 
    }
    \label{fig:ufomap_rendering}
\end{figure*}

% \subsection{Datasets}

% \subsection{Occlusions}

% \subsection{Semantic Segmentation}
% In this section we present a naive semantic segmentation implementation that makes use of UFOMap for information fusion. We present how we aggregate the output from an off the shelf semantic segmentation in the map and then query the map for the fused result. 

%\subsection{Occlusion calculations}
%UFOMap was designed to be a general mapping framework, but as we will demonstrate in this paper it well suited for many mapping related tasks in the autonomous driving setting. For scene understanding, in particular for risk assessment, it is important to be able to reason about occlusions. UFOMap uses an explicit representation of the unknown space. In contrast to, for example, OctoMap, voxels that have not yet been observed are create within the mapped volume. While this might seem wasteful, it is more than compensated for memory wise by using a more lean data structure. It also leads to great speedup when querying for unknown space. This allow us to query for space that has never been seen. Furthermore, what is currently occluded can be expressed as a query using the time step indicator.

\begin{table*}[!h] 
    \centering
    \caption{LiDAR based semantic segmentation evaluation using a map compared to single scan, in percent.}
    \newcommand*{\headformat}[1]{#1}
    \newlength{\maxlen}
    \settowidth{\maxlen}{\headformat{Other-vehicle}}
    \newcommand*{\head}[1]{%
        \begin{sideways}
            \makebox[\maxlen][l]{\headformat{#1}}
        \end{sideways}
    }
    \setlength{\tabcolsep}{4.0pt}
    \begin{tabular*}{\linewidth}{@{\extracolsep{\fill}}lc|ccccccccccccccccccc@{}} 
        \toprule
        
        & \head{mIoU} & \head{Car} & \head{Bicycle} & \head{Motorcycle} & \head{Truck} & \head{Other-vehicle} & \head{Person} & \head{Bicyclist} & \head{Motorcyclist} & \head{Road} & \head{Parking} & \head{Sidewalk} & \head{Other-ground} & \head{Building} & \head{Fence} & \head{Vegetation} & \head{Trunk} & \head{Terrain} & \head{Pole} & \head{Traffic sign} \\
        
        \midrule
        
        SPVNAS~\cite{tang2020searching}
        & 62.9 & 96.5	& 32.6 & 64.1 & 68.5 & 58.1 & 69.9 & 83.5 & 0.0 & 93.2 & 48.2 & \textbf{80.6} & 0.0 & 91.1 & 64.2 & 88.1 & 66.6 & 74.5 & 64.1 & 50.8 \\[0.1cm]
        UFOMap\(^\text{top-1}_\text{SPVNAS}\)%UFOMap\textsuperscript{top-1}
        %\textsuperscript{top-1} 
        & 64.1 & 96.0 & \textbf{39.5} & 67.7 & 82.9 & 66.4 & \textbf{74.0} & 83.2 & 0.0 & 91.8 & \textbf{48.8} & 78.2 & 0.0 & 91.1 & 62.3 & 87.3 & 63.5 & 72.0 & 62.6 & 50.1 \\[0.1cm]
        SPVCNN~\cite{tang2020searching}
        & 61.4 & 96.5	& 17.0 & 60.2 & 72.2 & 56.5 & 66.0 & 81.0 & 0.0 & \textbf{93.3} & 46.5 & 80.1 & 0.0 & 91.2 & 63.3 & \textbf{89.0} & 65.6 & \textbf{76.8} & 64.1 & 47.4 \\[0.1cm]
        
        UFOMap\(^\text{top-1}_\text{SPVCNN}\)%UFOMap\textsuperscript{top-1}
        %\textsuperscript{top-1} 
        & 62.7 & 96.0 & 18.6 & 64.8 & \textbf{86.9} & 66.5 & 71.3 & 82.3 & 0.0 & 92.2 & 45.5 & 78.0 & 0.0 & 91.4 & 62.0 & 88.5 & 63.1 & 75.4 & 62.7 & 45.6 \\[0.1cm]
        UFOMap\(^\text{top-1}_\text{fusion}\)
        %\textsuperscript{top-1} 
        & \textbf{65.1} & \textbf{96.8} & 28.6 & \textbf{69.4} & 85.5 & \textbf{70.4} & 73.7 & \textbf{85.3} & 0.0 & 92.9 & 48.5 & 80.0 & 0.0 & \textbf{92.2} & \textbf{67.0} & 88.6 & \textbf{67.6} & 75.4 & \textbf{64.5} & \textbf{50.9} \\[0.1cm]
        % UFOMap\(^\text{top-2}_\text{SPVNAS}\)%UFOMap\textsuperscript{top-2} 
        % & 75.8 & 98.7 & 53.8 & 83.0 & 94.8 & 84.9 & 80.9 & 89.0 & 0.0 & 97.3 & 63.0 & 90.4 & 0.0 & 96.2 & 86.2 & 95.2 & 85.8 & 88.2 & 82.3 & 70.0 \\

        \bottomrule
    \end{tabular*}
    \label{tab:semantic_container_comparison}
\end{table*}

\subsection{Map Manipulation and Runtime Comparison}
In this section, we show some qualitative results on how UFOMap can be used to manipulate map information.
We also compare its runtime performance with OctoMap~\cite{hornung2013octomap}.
% two widely used 3D mapping frameworks, namely Octomap~\cite{hornung2013octomap} and Voxblox~\cite{ol2016voxblox}. 

\subsection{Semantic Segmentation Using Information Fusion}
\label{subsec:setup_semseg}
% In this section, we demonstrate how UFOMap can be used to fuse single scan results from a LiDAR based semantic segmentation network, SPVNAS~\cite{tang2020searching}. Our choice of SPVNAS was motivated by its top performance on the semantic segmentation benchmark.
% % We chose an off the shelf neural network, SPVNAS~\cite{tang2020searching} in order to compute single scan semantic segmentation .
In this section, we evaluate the hypothesis that a 3D grid is beneficial for information fusion.
% mentioned in~\ref{sec:information_fusion}. 
% Concretely, we take an off-the-shelf approach for LiDAR-based semantic segmentation, namely SPVNAS and SPVCNN~\cite{tang2020searching}. Our choice was motivated by their top performance on the semantic segmentation benchmark. We fuse the single scan semantic estimates across time using UFOMap and evaluate if doing so helps improve the overall semantic segmentation.
Concretely, we take off-the-shelf DNN based approaches for LiDAR-based semantic segmentation and fuse their single scan semantic estimates across time using UFOMap. Thereafter, we evaluate if doing so helps improve the overall semantic segmentation.
For our baseline, we choose SPVNAS and SPVCNN~\cite{tang2020searching} neural networks, motivated by their top performance on the semantic segmentation benchmark.

Evaluation is carried out on the validation set (sequence \num{08}) of the SemanticKITTI dataset, containing \num{4071} scans. 
Following the official SemanticKITTI benchmark, we evaluate on \num{19} semantic classes and use the mean Jaccard Index shown in Equation \ref{eq:jacc}, also known as the mean intersection over union (IOU) as the metric~\cite{everingham2015pascal}.

\begin{equation}
    \frac{1}{C} \sum_{c=1}^{C}{\frac{TP_c}{TP_c+FP_c+FN_c}}
    % IoU(A,B) = \frac{|A \cap B|}{|A \cup B|}
\label{eq:jacc}
\end{equation}

here $TP_{c}$, $FP_{c}$ and $FN_{c}$ respectively are the true positive, false positive, and false negative estimates for semantic class \emph{c}, and \emph{C} is the total number of classes.

%The input is a point cloud containing semantic labels estimated by the neural network.
%The map is created with a resolution of \qty{10}{\centi\metre}. 
%In order to draw a fair comparison to a sample at time step \textit{t}, 

\looseness=-1
As estimates are accumulated temporally, a single voxel might encounter hits from multiple scans.
% \sout{As output at time step \(t\) we integrate the measurements (semantic estimates) until time step \(t\) into UFOMap and then extract the semantic labels with the highest confidence from the map.}
As output we first integrate the measurements (semantic estimates) until time step \(t\) into UFOMap. Then for each point at time step \(t\), we query the voxel that it belongs to for the label with highest confidence.
% and compare it to single scan result at the same time step.
Depending on the resolution of the map, the sequence, and the network we used, we observed that up to 15 $\%$ of voxels can have two or more labels with the same highest confidence. In these cases, we pick the label from the neural network to break the tie. We refer to this output as UFOMap\textsuperscript{top-1}.

To investigate the information fusion capabilities of the map, we also fuse the semantic estimates from SPVCNN and SPVNAS.
That is, each point in the input point cloud now contains two labels, corresponding to the top-1 labels of the two networks.
We refer to it as UFOMap\(^\text{top-1}_\text{fusion}\).
% \sout{We also present the \emph{top-2} results. That is, if any of the top-2 labels estimated by the map match the ground truth, we consider it as a correct estimate. We refer to this as UFOMap\textsuperscript{top-2}.}

\section{Experimental Results}
%\subsection{Map Manipulation}

\subsection{Semantic Segmentation Using Information Fusion}
\label{sec:res_semantics}

Table \ref{tab:semantic_container_comparison} shows the results from Section \ref{subsec:setup_semseg}.
We observe that accumulating information in UFOMap shows a marginal improvement in the mean IOU, when compared to single scan estimates from the networks.
Notably, fusing estimates from the two networks further improves the results, without the need for additional expensive labelling and neural network training.
% Usually the number of points for background far outweigh those of foreground.

%\sout{The IOU improves for foreground classes (movable objects), indicating that fusing the estimates is beneficial for classes which are statistically less represented in the point cloud. On the other hand, the IOU becomes worse for the background classes (static objects).
%% Usually the number of points for background far outweigh those of foreground. 
%This could be caused by the fact that the boundaries between different background classes (e.g., parking and drivable surfaces) are less well defined compared to the foreground classes;
%% \sout{hence, it is expected that the neural network will produce more noisy estimates on these boundaries, and fusing these estimates leads to a decrease in IOU.}
% hence, voxelization in these areas could lead to semantic mismatch. Decreasing the voxel size should help remedy this.}

The IOU improves for foreground classes (movable objects), indicating that fusing the estimates is beneficial for classes which are statistically less represented in the point cloud. On the other hand, the IOU becomes worse for the background classes (static objects). The foreground objects are more likely to be surrounded by free space; hence all the points in surrounding area are likely to have the same label. On the other hand, the background classes often share borders. Hence when querying in these regions, voxelization could lead to mismatch in semantics. Decreasing voxel size should reduce this mismatch, but at same time will lead to worsening performance for foreground objects, due to the reduced accumulation effect.

% \textcolor{red}{Considering many voxels have two or more labels with the same highest confidence, we also compute the top-2 mean IOU. That is, if any of the top-2 labels estimated by the map match the ground truth, we consider it as a correct estimate. This value is found to be 75.8\%  for UFOMap\textsubscript{SPVNAS}, showing} 
% % \sout{We also observe that UFOMap\textsuperscript{top-2} shows overall improvements across all classes. This shows that} 
Notably, even though we only integrate the top-1 label from the network in a voxel at each iteration, a distribution is still formed in the map. That is, each voxel can store multiple semantic labels. This suggests that better estimates can be extracted by, for example, incorporating geometric information or clustering and smoothing techniques.

% This is not unexpected.
% But notably, the top-2 estimate of the map is 
% derived from the best (top-1) estimate of the neural network. 

% \subsection{Manipulate map data for downstream tasks}
\subsection{Map Manipulation and Runtime Comparison}
% UFOMap allows for efficient querying of the information using a set of predicates.  %The framework supports assigning readable strings to the  labels. For example if labels 17000-17999 are known to contain car like objects, one could assign them a string label ``\textit{car}''.
% %The framework also includes predicates which enable fast voxel search. % 
% %The predicates could be combined to generate powerful queries.
% It is possible to make queries both connected to the spatial location and content. 
% %To bound the search further, the framework also provides bounding volume predicates.
% One could, for instance, look for all ``cars'' or ``pedestrians'' within a 100m radius of the ego vehicle. 

In the top row of Figure \ref{fig:ufomap_rendering} we see the color and semantic information for a section of the map. We can efficiently extract, for example, background/foreground objects (bottom row).
Using queries based on semantics, one can easily obtain representations tailored to different use cases. For instance, similar to~\cite{poggenhans2018lanelet2}, one can retain the observable elements in the surroundings (e.g. buildings, pole-like objects, traffic lights), to obtain a map usable for localization.  

\begin{figure}[!t]
    \centering
%    \scalebarandzoomclip{1.0\linewidth}{figures/full_time_step.png}{0.4,0.25}{0.5,0.35}{0.97\linewidth}{figures/new_old.png}
    \includegraphics[width=0.97\linewidth,trim={0cm 0cm 0cm 5cm},clip]{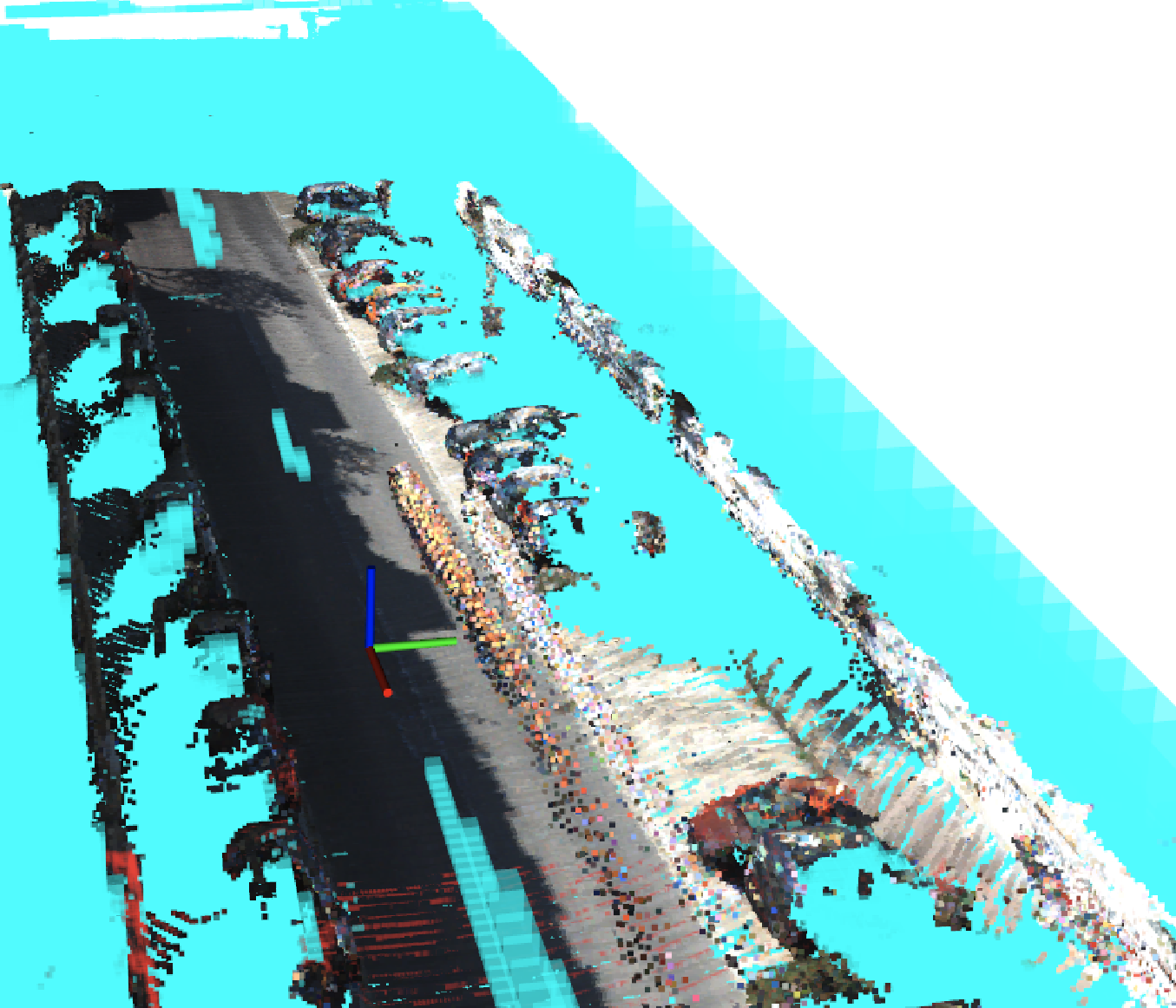}
    \caption{%Top: Same as~\protect\ref{fig:cool_figure}, colors show time step stored in voxel. Blue to red imply older to newer time steps. Bottom: 
    \looseness=-1
    Showing part of map while building it. The axes represent the ego vehicle, traveling upwards. The parts marked in cyan are either old (based on the time step stored in the voxels) or unknown information. As can be seen, space behind the ego vehicle is classified as unknown again.}
    \label{fig:ufomap_unknown}
\end{figure}

Occlusions play an important role in assessing risk for autonomous vehicles. UFOMap provides mechanisms for finding such regions. 
% \ref{fig:ufomap_time_step} illustrates the time-step indicator information in the map. Using this,
Using the time step indicator, as shown in Figure~\ref{fig:ufomap_time_step},
one can look at voxels that have not been updated since a certain time. Furthermore, we can include the cells that are in the unknown state. Figure \ref{fig:ufomap_unknown} %illustrates the time information in the map using colors (top part). The time goes from blue to red. The lower part 
shows a snapshot of a scene where all voxels that are occluded are marked in cyan. 

% The computation time details are listed in~\ref{tab:computational_time}.
Figure~\ref{tab:computational_time} lists the computational performance averaged over a sequence containing 4540 scans.
% for both OctoMap and UFOMap.
UFOMap far exceeds OctoMap in terms of 
the memory efficiency, 
the time taken to integrate measurements into the map, as well as publishing the map to distribute it for use by different components.
OctoMap publishes the entire map at every iteration, making it prohibitively slow. For this experiment, the publishing rate for OctoMap was reduced to once every hundred iterations.
% Notably for this experiment, the publishing rate for OctoMap was reduced to once every hundred iterations. This is because Octomap publishes the entire map every iteration, making it prohibitively slow.
In comparison, UFOMap publishes only the updated part of the map, making the publishing at each iteration scalable.
% In comparison, publishing only the updated part of the map at each iteration allows UFOMap to be scalable.
Notably at any time, the end result remains the same at the receiving end, i.e., the entire map of the environment.

\begin{table}[!b]
    \centering
    \caption{Single threaded performance on sequence 00.}
    \sisetup{table-align-uncertainty,table-format=1.2(3)}%
    \setlength{\tabcolsep}{5.2pt}%
    \begin{tabular}{@{}lccSS[table-format=2.2(3)]S@{}}
        \toprule
        \bfseries Map &
        \thead{\bfseries Res.\\(cm)} & \thead{\bfseries Mem.\\(GiB)} & {\thead{\bfseries Integrate\\(s)}} & {\thead{\bfseries Publish\\(s)}} & {\thead{\bfseries Query\\(s)}} \\
        
        \midrule
        UFOMap & \multirow{2}{*}{10} & 1.14 & 0.05(1) & 0.04(1) & 0.07(4) \\
        OctoMap &   & 9.46 & 1.62(31) & 20.65(1095) & 1.75(28) \\[0.1cm]
        UFOMap & \multirow{2}{*}{20}  & 0.43 & 0.03(1)  & 0.01(1)   & 0.06(3) \\
        OctoMap &   & 1.48 & 0.41(10) & 3.49(185) & 1.69(25) \\
        
        \bottomrule
    \end{tabular}
    \label{tab:computational_time}
\end{table}

Furthermore, we query for all unknown voxels at 20 cm resolution within a 200 x 200 x 10 $m^3$ axis aligned bounding box around the vehicle.
% For 10 cm resolution we choose query one level up, and leaf level for 20 cm resolution
% The query resolution chosen is \qty{20}{\centi\metre}
For UFOMap we can directly query for the unknown voxels, or those which haven't been updated within the last 5 seconds, leading to a significant speedup.
% Also, by exploiting the hierarchical nature of the map, queries can be speeded up significantly. 
Exploiting the hierarchical nature of the map can also be beneficial. 
For example, for 10 cm resolution we query for unknown space one level up the tree, and for 20 cm resolution we do this at leaf level; hence, in both cases we effectively query for information at 20 cm. 

\section{Summary and Conclusions}
\label{sec:concl}
%summary 
In this paper, we provide a comprehensive review of existing map representations for autonomous driving. Furthermore, we introduced UFOMap as a way to achieve real-time semantic 3D mapping for outdoor environments. % Based on that we find that 
% online dense 3D mapping is not widely used in driving scenarios, mainly due to lack of efficient implementations.
%existing implementations are not sufficient for this task.

% limitation and future work 
% Although we showed that the UFOMap framework outperforms some of the state-of-the-art methods,
% Although we showed that the UFOMap framework meets many requirements which we believe are needed for advanced vehicle autonomy, there is still a lot to do.
At the moment, UFOMap does not handle dynamic objects well. This can be observed in Figure~\ref{fig:ufomap_unknown}, where two cyclists enter the scene, thus inducing noise. Incorporating the dynamic behavior of the environment into our framework is the next step in the research. 
%Additionally, UFOMap does not have localization abilities. 
%We think localization is an essential part of autonomous driving and we will try to add it in the next version. 
Finally, as we described in Section~\ref{sec:res_semantics} there are several directions to investigate to improve the semantic segmentation results.
% One could consider using a better integration model, incorporating the full distribution of 
Instead of using the best estimate provided by the neural network, 
one could consider incorporating the distribution over all semantic classes, using an integration model similar to~\cite{Rosinol20icra-Kimera}.
%&we have to work on the results aggregation to successfully improve the confidence of the true label and be able to achieve the improvement of $15\%$ for UFOMap\textsuperscript{top-1}.   
Fusing estimates from multiple sensor modalities for semantic segmentation networks should 
% further improve the results.
% lead to more improvements.
also prove beneficial.

% Serve as an intermediate representation between sensors and other more dedicated map representations.
% * Rely less on prior map information.
% * Can provide the ground height information from rasterized HD Maps.

% - show that it may be also beneficial for the outdoor
% - abrv of the conference names 
% - skip pages 
% - just 1st person and 1st letter for the author
% \bibliography{ref}

% \looseness=-1
\balance
% \addtolength{\textheight}{-5cm}
\printbibliography

\end{document}